\documentclass[letterpaper]{article} 
\usepackage{aaai25}  
\usepackage{times}  
\usepackage{helvet}  
\usepackage{courier}  
\usepackage[hyphens]{url}  
\usepackage{graphicx} 
\usepackage{natbib}  
\usepackage{caption} 
\usepackage{algorithm}
\usepackage{amssymb}
\usepackage{amsmath} 
\usepackage{multirow}
\usepackage{booktabs}
\usepackage[table]{xcolor}

\usepackage{algorithmicx}
\usepackage{makecell}

\usepackage{colortbl}
\usepackage{xcolor}
\usepackage{amsmath}
\usepackage{multirow}
\usepackage{graphicx}
\usepackage{adjustbox}

\newcommand{\eg}{\textit{e.g., }}
\newcommand{\ie}{\textit{i.e., }}

\usepackage{color}

\usepackage{newfloat}
\usepackage{listings}
\DeclareCaptionStyle{ruled}{labelfont=normalfont,labelsep=colon,strut=off} 
\floatstyle{ruled}
\newfloat{listing}{tb}{lst}{}
\floatname{listing}{Listing}
\title{MaDis-Stereo: Enhanced Stereo Matching via Distilled Masked Image Modeling}
\author{
    Jihye Ahn\textsuperscript{\rm 1},
    Hyesong Choi\textsuperscript{\rm 1},
    Soomin Kim\textsuperscript{\rm 1},
    Dongbo Min\textsuperscript{\rm 1, *}
}
\affiliations{
    \textsuperscript{\rm 1}Ewha W. University\\
    ajh0531@ewha.ac.kr, hyesong@ewha.ac.kr, kim.soomin@ewha.ac.kr, dbmin@ewha.ac.kr

}

\begin{document}

\maketitle

\begin{abstract}
{
In stereo matching, CNNs have traditionally served as the predominant architectures. Although Transformer-based stereo models have been studied recently, their performance still lags behind CNN-based stereo models due to the inherent data scarcity issue in the stereo matching task. In this paper, we propose \textbf{Ma}sked Image Modeling \textbf{Dis}tilled \textbf{Stereo} matching model, termed \textbf{MaDis-Stereo}, that enhances locality inductive bias by leveraging Masked Image Modeling (MIM) in training Transformer-based stereo model. Given randomly masked stereo images as inputs, our method attempts to conduct both image reconstruction and depth prediction tasks. While this strategy is beneficial to resolving the data scarcity issue, the dual challenge of reconstructing masked tokens and subsequently performing stereo matching poses significant challenges, particularly in terms of training stability. To address this, we propose to use an auxiliary network (teacher), updated via Exponential Moving Average (EMA), along with the original stereo model (student), where teacher predictions serve as pseudo supervisory signals to effectively distill knowledge into the student model. State-of-the-arts performance is achieved with the proposed method on several stereo matching such as ETH3D and KITTI 2015. Additionally, to demonstrate that our model effectively leverages locality inductive bias, we provide the attention distance measurement.
}
\end{abstract}

\section{Introduction}

Stereo depth estimation is a critical task in computer vision, focused on predicting disparities between stereo image pairs. Its importance is widely recognized across various applications, including autonomous driving~\cite{deepdriving, stereodriving}, SLAM~\cite{slamalgo, slamov}, robotic control~\cite{stereorobot}, drone navigation~\cite{drone}, and beyond. Recently, the performance of the stereo matching has been improved remarkably by adopting deep neural networks (DNNs), but the difficulty of collecting large-scale ground truth training data using costly equipment such as LiDAR~\cite{LiDAR1, LiDAR2} becomes a factor that hinders the advance of the stereo matching task.

For this reason, the stereo models based on Convolutional Neural Networks (CNNs)~\cite{stereoreview} are still a widely used solution for stereo matching tasks, unlike other vision tasks where the Transformer architectures~\cite{vit, swin} are gradually becoming mainstream. In stereo matching, the CNNs-based models outperform Transformer-based counterparts~\cite{sttr, chitransformer, unifying} due to the training data efficiency resulting from the local inductive bias inherent in convolutional operations.

This phenomenon contrasts with the prevailing trends observed in most computer vision tasks. Since the introduction of ViT~\cite{vit}, Transformer-based approaches have been recognized for their capacity to effectively learn global representations from large-scale datasets, achieving state-of-the-art results across numerous vision applications such as image classification~\cite{classificationT2, classificationT3, classificationT1}, 2D/3D object detection~\cite{obj1, obj2, obj3}, and semantic segmentation~\cite{seman2, seman3, seman4, seman1}. 
Since the Transformer typically requires extensive training data to obtain overwhelming performance, the currently limited ground truth data for stereo matching is an obstacle to achieving competitive performance in the Transformer-based stereo matching networks.

\begin{figure}[t]
\center
\includegraphics[width=0.9\columnwidth]{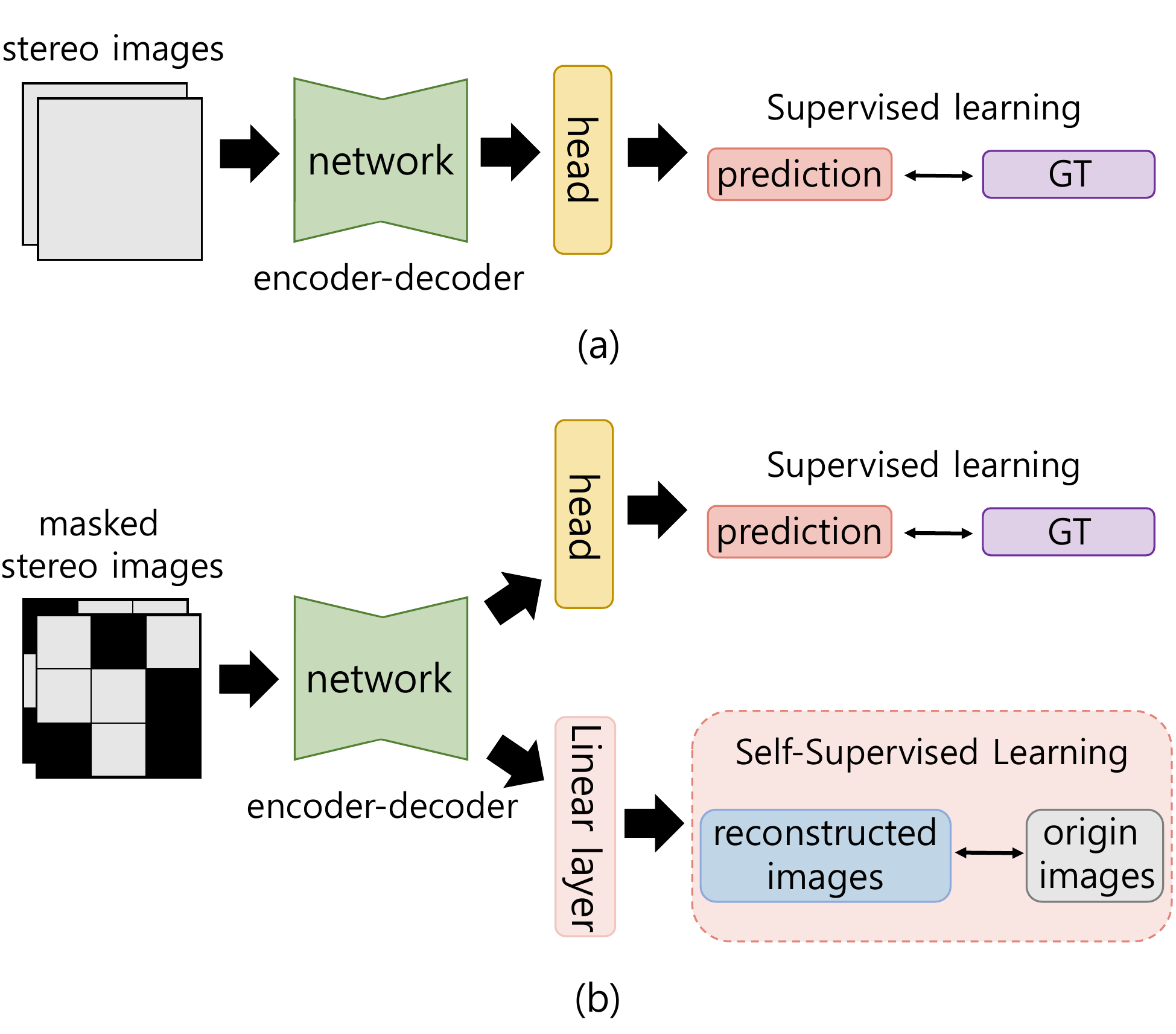}
\caption{{\bf Comparative Illustration of Conventional Approaches}; (a) Current supervised learning approaches for stereo depth estimation, either implemented with CNNs~\cite{raftnet, crestereo} or Transformer~\cite{croco2}, typically train the model with a pre-trained encoder by comparing predicted disparity maps with ground truth depth labels using a supervised depth loss. (b) Initializing with the Transformer encoder pre-trained using Masked Image Modeling (MIM)~\cite{MAE}, the proposed method leverages both the self-supervised masking-and-reconstruction strategy and the supervised depth loss for stereo depth estimation.}
\label{fig:concept}
\vspace{-0.3cm}
\end{figure}

Nonetheless, considering the potential of the ever-evolving Transformer architecture, a few attempts have been made to overcome the above-mentioned limitations of Transformer-based stereo matching networks.
CroCo-Stereo~\cite{croco2} partially alleviates the data scarcity issue by leveraging the pre-trained encoder as a backbone. The pre-trained encoder~\cite{croco} is tailored for geometric downstream tasks by performing cross-view completion for massive stereo image pairs. While this approach addresses the data scarcity issue to some extent through the introduction of the pre-training methodology based on cross-view completion, Transformer models—designed primarily for learning global representations with extensive training data—continue to struggle with the complexities of stereo matching tasks.
In the fine-tuning stage for stereo matching, a sufficient amount of training data with ground truth labeled data is still necessary.

To address this data scarcity issue, it is crucial to incorporate locality inductive bias when fine-tuning the whole stereo model, in addition to pre-training the Transformer encoder. Recently, it was reported in~\cite{revealing} that Masked Image Modeling (MIM), which tends to concentrate around pixels in attention heads, introduces locality inductive bias in pre-trained Transformer models. The effectiveness of this locality is closely related to the masking ratio and masked patch size during pre-training.

Inspired by the benefits from MIM~\cite{revealing}, we propose a novel approach, termed \textbf{Ma}sked Image Modeling \textbf{Dis}tilled \textbf{Stereo} matching model (\textbf{MaDis-Stereo}), which is a Transformer-based model that incorporates MIM~\cite{igpt, ibot, simmim, MAE, siamese} that aims to impart locality inductive bias to the stereo depth estimation. This is achieved by pre-processing the stereo images through random masking and subsequently reconstructing the masked regions of images within the stereo matching network as described in Fig.~\ref{fig:concept}.

However, the straightforward application of the MIM without appropriate measures may not suit fine-tuning tasks such as stereo depth estimation well. This challenge stems from the discrepancy between the pre-training task, where MIM is utilized, and the subsequent fine-tuning tasks. In the MIM based pre-training, the primary objective is to mask and reconstruct the image. In contrast, when applying the MIM strategy directly to the stereo depth estimation model, the training process becomes more complex. In addition to the reconstruction of masked patches, it must predict disparities from the reconstructed patches, posing additional challenges beyond standard MIM pretraining. To mitigate the difficulty of these challenging tasks and facilitate effective and robust training, we propose to use additional supervision from teacher networks. Additionally, the masking ratio is set at 40\%, which is lower than the standard used in MIM pre-training. This adjustment was determined experimentally to establish an optimal upper bound, ensuring the model can effectively handle the dual challenge of reconstructing masked tokens and subsequently performing stereo matching.

Experimental validation on KITTI 2015~\cite{kitti15} and ETH3D~\cite{eth3d} datasets confirms that our model exhibits improved performance compared to the existing stereo depth estimation approaches.
Our contributions can be summarized as follows:
\begin{itemize}
    \item We introduce a novel stereo depth estimation framework that leverages Masked Image Modeling (MIM) applied to stereo images to obtain a locality inductive bias, which is particularly suitable for dense prediction tasks.
    \item We empirically demonstrate that the effective utilization of local image representation learning techniques derived from self-supervised learning significantly enhances the stereo matching process.
    \item We present a method to employ pseudo disparity maps generated by an Exponential Moving Average (EMA) teacher as supplementary guidance, thereby improving both stability and performance in disparity estimation.
\end{itemize}

\section{Related Work}

\subsubsection{Stereo Depth Estimation}
Stereo depth estimation is a fundamental task that involves determining the disparity of objects within a scene using stereo images, simulating the depth perception capabilities of the human visual system ~\cite{chitransformer}. Stereo depth estimation is widely utilized in various fields, including autonomous driving~\cite{li2019stereo, wang2019pseudo, chen2020dsgn}, robotics~\cite{wang2023application, nalpantidis2010stereo}, and 3D scene reconstruction~\cite{ju2023dg}, where accurate depth information is crucial for navigation, object detection, and environmental interaction. Moreover, it is increasingly being used as ground truth labels in monocular depth estimation tasks~\cite{tonioni2019unsupervised, choi2021adaptive}, further enhancing the training and accuracy of monocular models. With the advancement of transformer frameworks~\cite{vit, swin, chen2021pre, ranftl2021vision, lee2022knn, lee2023cross}, stereo-matching models have evolved significantly, offering enhanced capabilities in stereo depth estimation tasks. Stereo matching models can be broadly categorized into CNN-based and Transformer-based frameworks. Historically, most research has focused on CNN-based models, which construct a 3D cost volume~\cite{3dvol1, 3dvol2, 3dvol3} to calculate the disparity between corresponding pixels from the stereo images. An alternative approach concatenates features from both images to create a 4D cost volume~\cite{4dvol1, psmnet}, followed by cost aggregation to refine the volumes. In contrast, Transformer-based models~\cite{croco2, chitransformer} employ a different approach. These models utilize a cross-attention layer to facilitate the transfer of information across varying views~\cite{crossatten}, thereby eliminating the need for generating a correlation volume. Several studies~\cite{corres1, corres2} demonstrated that the positional encoding inherent in transformers helps establish spatial correspondence between different views. However, these Transformer-based models require massive ground truth training data to achieve competitive performance. Thus, there are fewer Transformer-based models than the latest CNN-based concurrent models~\cite{mocha, adaptive, selective, mcstereo}.

\subsubsection{Self-Supervised Learning}
Self-Supervised Learning (SSL) is designed to extract meaningful representations from large amounts of unlabeled data~\cite{pretext}. This approach facilitates the fine-tuning of models for various downstream tasks. At the core of SSL are carefully crafted pretext tasks that leverage the intrinsic patterns and relationships within the data. These tasks often involve intentionally removing specific image features, such as color~\cite{colorization}, followed by training the model to reconstruct the missing details. Broadly, pretext tasks fall into two categories: (1) augmentation-based~\cite{aug1, aug2} and (2) reconstruction-based methods~\cite{ssl}. Augmentation-based methods generate semantically similar outcomes through various transformations, while reconstruction-based methods focus on reconstructing hidden patches or pixels to discern object structures within the image. Despite the absence of explicit semantic supervision, networks trained on these pretext tasks show their effectiveness in learning rich and meaningful image representations from the data. This capability to use large-scale unlabeled datasets has garnered significant attention, particularly in tasks where the availability of labeled data is constrained.

\subsubsection{Mask Image Modeling}
Motivated by the successes in natural language processing (NLP) exemplified by BERT~\cite{bert} and the advent of ViT~\cite{vit}, a variety of self-supervised pre-training methodologies using Masked Image Modeling (MIM) have emerged. These approaches draw conceptual parallels to denoising autoencoders and context encoders, with the primary objective of reconstructing masked pixels, discrete tokens, or deep features. Initial efforts, such as iGPT~\cite{igpt}, which focused on pixel reconstruction, and ViT~\cite{vit}, which aimed to predict the average color of masked patches, fell short of achieving competitive performance relative to supervised models. However, the introduction of BEiT~\cite{beit}, which predicts visual tokens derived from a pre-trained Variational Autoencoder (VAE)~\cite{VAE}, marked a milestone in the evolution of self-supervised learning. Further notable advancements include the Masked Autoencoder (MAE)~\cite{MAE}, which employs raw pixel prediction for pre-training and emphasizes the importance of a high mask ratio (\eg, 75\%) due to spatial redundancy inherent in images. The MultiMAE~\cite{multimae} framework has been developed to extend its applicability to multi-modal or multitask scenarios. Furthermore, lightweight approaches~\cite{esam} are being developed to mitigate the substantial training resource requirements of MIM models. Recently, MTO~\cite{mim2} optimizes masked tokens to significantly enhance pre-training efficiency, while SBAM~\cite{mim1} introduces a saliency-based adaptive masking strategy that further refines the process by dynamically adjusting masking ratios based on token salience. 

\begin{figure*}[t]
\centering
\includegraphics[width=1.0\textwidth]{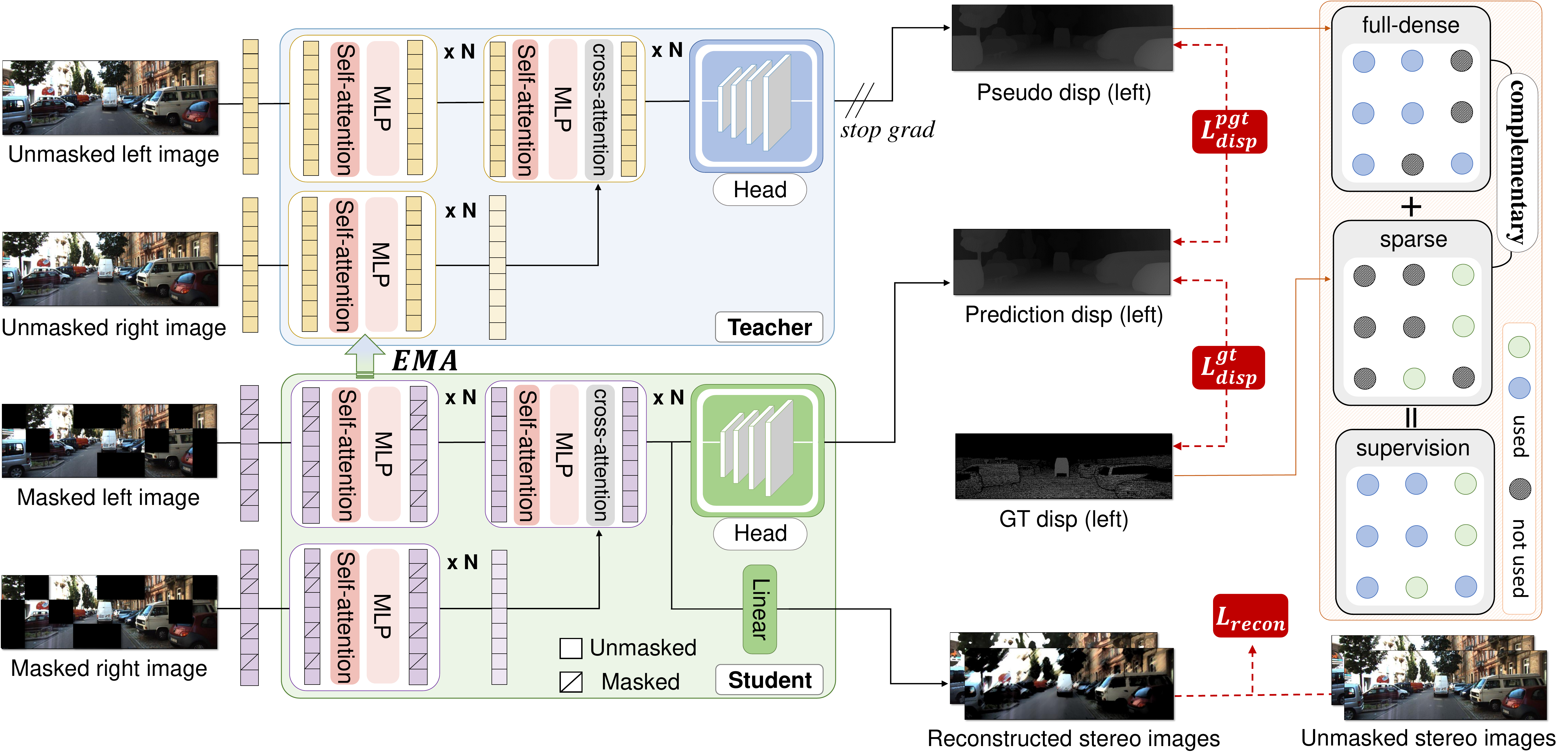} 
\caption{{\bf Overall architecture of the proposed MaDis-Stereo}. { It consists of (bottom) a main stereo network (student) processing masked stereo images and (top) an auxiliary stereo network (teacher), whose weight parameters are updated with exponential moving average (EMA). 
The ViT-Base~\cite{vit} encoder processes visible tokens from both the masked left and right views to extract image features. The left and right features are then fed into the ViT-Base decoder consisting of cross-attention blocks. Following~\cite{croco2}, the RefineNet-based feature fusion block~\cite{refinenet} is employed as a head module to produce disparity maps, and an additional linear layer is used to reconstruct the masked image patches~\cite{simmim}. Here, $N$ is set to 12.} The ground truth disparity maps and pseudo disparity maps are utilized as supervision signals in a complementary manner, with the blue and green circles on the right representing the pseudo disparities (`full-dense') generated by the teacher network and the ground truth disparities (`sparse'), respectively.}
\label{fig:architecture}
\end{figure*}

\section{Proposed Method}
\subsection{Motivation and Overview}
Although Transformer-based models exhibit promising performance, they typically require a large scale of training data compared to CNN-based models. The Masked Image Modeling (MIM) approach offers a potential solution to this data scarcity issue by providing the locality inductive bias into the model. Motivated by the strengths of MIM, we present a novel stereo depth estimation framework, referred to as MaDis-Stereo. However, the straightforward integration of MIM complicates the dual tasks of image reconstruction and disparity prediction, which can introduce model instability. To circumvent these issues, we introduce a novel strategy that effectively integrates MIM into the model while maintaining its stability.

Fig.~\ref{fig:architecture} illustrates the overall architecture of the proposed method. The MaDis-Stereo comprises a teacher network and a student network. The student network is constructed with a ViT-Base~\cite{vit} encoder to extract features from visible tokens in given images. The ViT-Base decoder consists of plain Transformer decoder blocks including self-attention among token features from the left image, cross-attention with token features from the right image, and a linear layer. A predicted disparity is generated by passing features gathered from different intermediate decoder blocks into the head module~\cite{refinenet}. In addition, the reconstructed image is produced by a prediction layer. The teacher network shares the same architecture as the student network except a prediction layer for reconstruction, since it takes unmasked images as inputs. During training, the parameters of the teacher network are updated via exponential moving average (EMA) from the student network. The student network is trained using both the ground truth disparities and pseudo disparity maps generated by the teacher network.

\subsection{Masking-and-Reconstruction of Stereo Images} \label{Sec:mim}
MaDis-Stereo's key architectural innovation lies in its capability to learn local representations through a masking-and-reconstruction methodology. For that, we first divide given left and right images $I_l$ and $I_r$ into $N$ non-overlapping patches represented as $ I_{l}=\left\{I_{l,1},..., I_{l, N} \right\}$ and $ I_{r}=\left\{I_{r,1},..., I_{r,N} \right\}$, and then apply random masking to them. A masking ratio $\beta\in \left [ 0,1 \right ]$ is provided to the model as a hyperparameter, which determines the proportion of masked areas within input data. Simply speaking, $n$ patches are masked as follows; $n = \beta \cdot N$.

A set of visible left patches is indicated by $\tilde{I_l}=\left\{I_{l,i}|m_i=0 \right\}$, where $m_i=0$ signifies that the patch $I_{l,i}$ is not masked, and $m_i=1$ means otherwise. The same random masking strategy is also applied to the right image. MaDis-Stereo proceeds with the reconstruction process using the masked view as input thereafter. $\tilde{I_l}$ and $\tilde{I_r}$ are processed individually by the ViT~\cite{vit} encoder $E_\theta$. Our model reconstructs the image $\hat{I_l}$ using a linear layer after decoding the image features with $D_{\phi}$. Here, $\theta$ and $\phi$ are the weight parameters of the Transformer encoder and decoder, respectively. In the decoder, MaDis-Stereo uses cross-attention to reconstruct the image. For reconstructing the left image, the left encoded feature $ E_\theta (\tilde{I_l})$ is used as a query, while the right encoded feature $ E_\theta (\tilde{I_r})$ is employed as a key and a value for allowing information exchange between the two views. The process of reconstructing the right image is conducted similarly, except for reversing the query $E_\theta(\tilde{I_r})$ and the key and value $E_\theta(\tilde{I_l})$. Formally, the left image reconstruction can be written as
\begin{equation}
  \hat{I_l}= {\mathtt{Linear}}\left( D_{\phi} ( E_\theta  ( \tilde{I_l} );E_\theta ( \tilde{I_r} ) )\right).
  \label{eq:mim} 
\end{equation} 

Simultaneously reconstructing masked images and estimating depth is a challenging task and can destabilize the network. Therefore, we use a relatively lower value of $r$ than the masking ratio typically used during the pre-training phase, \eg 60\% default masking ratio for SimMIM~\cite{simmim}. Experimentally, we found an appropriate masking ratio for MaDis-Stereo to be 40\%. 

\begin{figure}[t]
\center
\includegraphics[width=0.9\columnwidth]{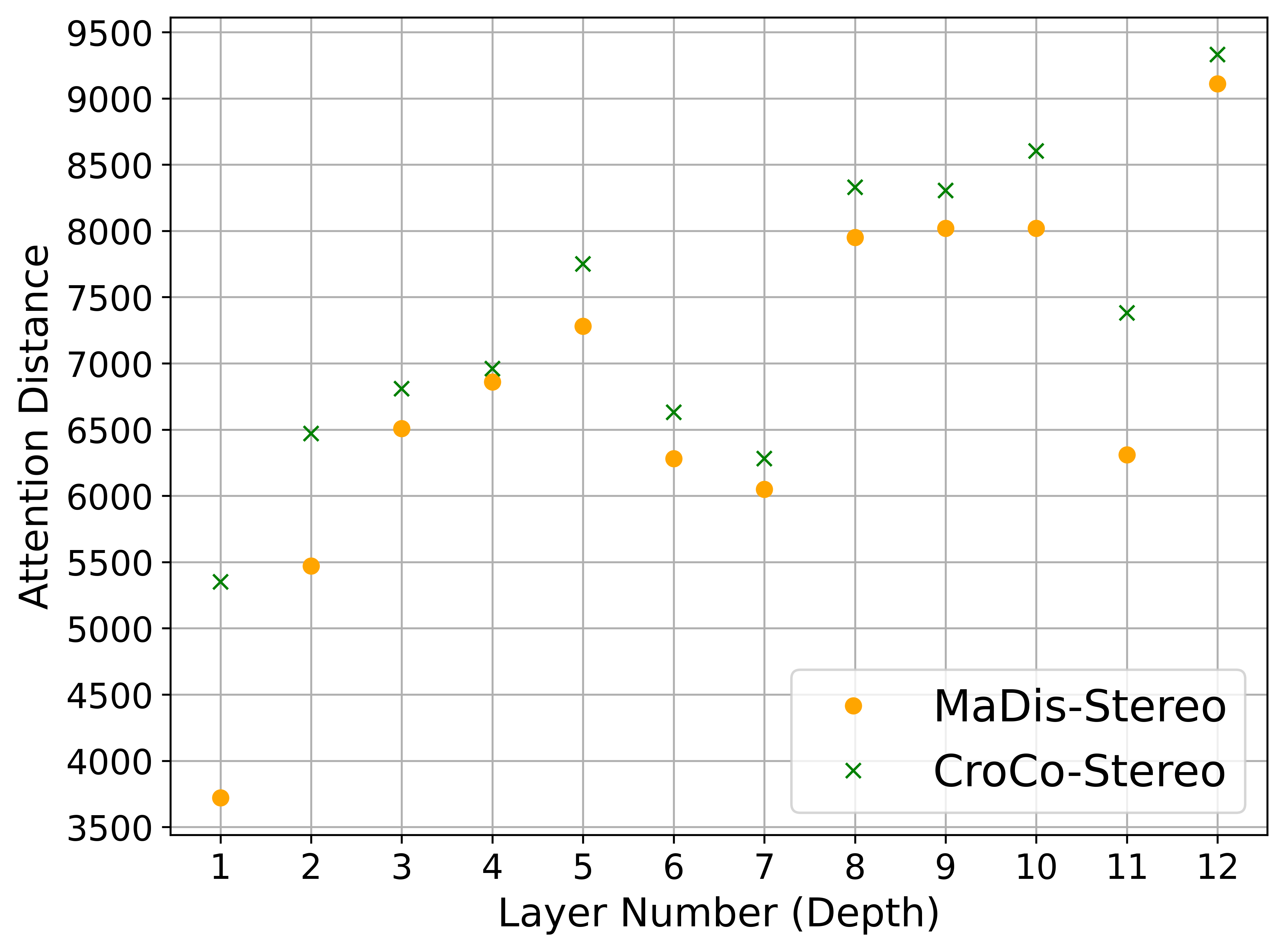}
\caption{{\bf Comparison of Attention Distance Maps}; It shows the result of computing averaged attention distance in 12 attention heads of each layer. The average attention distance~\cite{revealing} across various attention heads represented by dots indicates that MaDis-Stereo tends to focus locally compared to CroCo-Stereo~\cite{croco2}.}
\label{fig:attentionDistance}
\vspace{-0.3cm}
\end{figure}

Additionally, Fig.~\ref{fig:attentionDistance} demonstrates that our approach enhances the model's capability to focus more effectively on local patterns within the images. The attention distances were computed by obtaining attention weights from the cross-attention layer after processing the KITTI 2015~\cite{kitti15} dataset through the model. The comparison is made between our model, which employs a masking-and-reconstruction strategy, and a standard Transformer-based stereo matching model~\cite{croco2}. A small attention distance indicates that the model's attention heads focus more on nearby pixels, demonstrating stronger locality inductive bias.

\subsection{Network Architecture and Training Procedure}

As illustrated in Fig.~\ref{fig:architecture}, the architecture comprises two networks: (1) the student network and (2) the teacher network updated via Exponential Moving Average (EMA). Both networks are built upon the ViT~\cite{vit} encoder for feature extraction, the Transformer decoder with multi-head cross-attention layers, and the RefineNet-based feature fusion block~\cite{refinenet} as a head module for disparity prediction. Additionally, the student network includes a prediction module composed of a linear layer for reconstructing masked image patches, following the methodology outlined in SimMIM~\cite{simmim}.

The overall training procedure of MaDis-Stereo is described as follows. We provide the masked stereo images obtained by applying random masking and original stereo images without masking as inputs. The two masked views are fed into the student network and are then reconstructed to impose the locality inductive bias. Note that this learning process is only conducted within the student network. In the student network, the feature of the masked image is extracted through the ViT~\cite{vit} encoder by processing visible patches, and then the ViT~\cite{vit} decoder uses the multi-head cross-attention layers to facilitate information exchange between the left and right features. The decoded features are subsequently passed through a linear layer for image reconstruction, while also being input into a RefineNet-based feature fusion block~\cite{refinenet} to predict the output disparity. The RefineNet-based fusion block generates the disparity map by reshaping and merging four features from several transformer decoder blocks using convolutional layers.

On the other hand, the teacher network does not undergo a process of reconstructing masked regions since intact stereo images are used as inputs and only perform disparity prediction. While the student network continues its learning process, the teacher network maintains a stop-gradient state. The encoder of the teacher network is EMA-updated using the encoder of the student network, ensuring smooth training. The overall structure follows the framework of self-supervised learning methods~\cite{grill2020bootstrap, chen2021exploring, ema1, ema2}.
In our model, the EMA update is applied up to the teacher ViT~\cite{vit} encoder, and the Transformer decoder and feature fusion block~\cite{refinenet} are built in a Siamese manner. Formally, the teacher encoder parameter $\theta^T_{t+1}$ at iteration $t+1$ is EMA-updated using the student encoder parameter $\theta_{t+1}$ as 
\begin{equation}
  \theta^T_{t+1}\leftarrow \alpha \theta^T_{t}+(1-\alpha)\theta_{t+1},
  \label{eq:eq0} 
\end{equation}
\noindent where $\alpha$ is a hyperparameter of EMA update, set to 0.9999 in our work.

\subsection{Distilling Teacher Knowledge to Student Model} \label{Distilled}
Performing image reconstruction and depth estimation simultaneously can be a challenging task. To alleviate the challenges of the task, we utilize disparity maps generated by the teacher network as supplementary pseudo-labels to guide the student network. To be specific, the ground truth disparity maps are typically sparse due to the inherent limitation of active depth sensors~\cite{LiDAR2} used to collect the training data.
Employing pseudo labels facilitates effective model training in scenarios of sparse ground truth disparity maps. These dense pseudo disparity maps act as supervisory signals for regions where ground truth disparity values obtained by LiDAR are unavailable.

From another perspective, using the pseudo labels of the teacher network facilitates effective knowledge transfer to the student network, ultimately enhancing the performance of the student network. Similar approaches have also been employed in weakly/unsupervised domain adaptation~\cite{weakly, DApseudo}, where the model is trained using labeled source data and is then applied to the target domain to generate pseudo labels using the teacher network. It was also reported that these pseudo labels help train the student model.

\subsection{Loss Function}

\subsubsection{Disparity loss.}
MaDis-Stereo parameterizes the network output using a Laplacian distribution~\cite{lap}, similar to CroCo-Stereo~\cite{croco2}. Given a stereo image input, the model predicts a disparity map $d$ and a scale parameter $\sigma$. The training objective involves minimizing the negative log-likelihood function of the ground-truth target disparity map $d^{gt}$ and pseudo-disparity map $d^{pgt}$ computed from the EMA-updated teacher model as follows:
\begin{align}
  L_{disp} &= \;  L^{gt}_{disp} + L^{pgt}_{disp} \nonumber\\
  &= \;\frac{1}{|\Omega|}  \left[ \sum_{i\in G}\left( \frac{\left| d_{i}-d^{gt}_{i}\right|}{\sigma_{i}} - 2\log \sigma_{i} \right) \right. \nonumber\\
  &\quad + \left. \sum_{j\in \Omega \setminus G}\left( \frac{\left| d_{j}-d^{pgt}_{j}\right|}{\sigma_{j}} - 2\log \sigma_{j} \right) \right],
  \label{eq:eq2}
\end{align}

\noindent where $L^{gt}_{disp}$ and $L^{pgt}_{disp}$ are loss functions defined using $d^{gt}$ and $d^{pgt}$, respectively. 
The scale parameter $\sigma$ serves as an indication of predictive uncertainty: higher values of $\sigma$ result in less stringent penalties for larger errors, while lower values of $\sigma$ lead to greater rewards for accurate predictions.

$\Omega$ and $G$ indicate a set of all pixels in the image and a set of pixels where ground truth depth labels are available, respectively. 
Since the ground truth depth labels $d^{gt}$ are typically sparse, we use the dense pseudo labels $d^{pgt}$ generated by the teacher network as an additional guidance.
Due to the potential inaccuracies of the pseudo labels $d^{pgt}$, we exclude $d^{pgt}$ for pixels where the $d^{gt}$ is available. Specifically, for pixel $i$ where ground truth exists, the training process relies solely on the $d^{gt}_i$, avoiding any guidance from the pseudo labels $d^{pgt}_i$. For the location of pixel $j$ where ground truth $d^{gt}_j$ is unavailable (\ie, where $j \in \Omega \setminus G$), the pseudo labels $d^{pgt}_j$ can serve as a guidance. 

Similar to the semi-supervised learning paradigms~\cite{pseudolabel, fixmatch}, the disparity maps generated from the EMA-teacher model are used as the pseudo depth labels. Specifically, the student network takes masked images with incomplete information as inputs and predicts disparity maps with lower accuracy than the teacher. In contrast, the teacher network predicts pseudo disparity maps from unmasked (original) images, which are more precise compared to those from the student network.

\subsubsection{Image Reconstruction loss.}
The reconstruction loss is computed by predicting the pixels within masked regions of the input and evaluating $L_1$ loss, $L_{img}$, between the reconstructed images and the original ones.
$L_{img}$ is computed for the masked patches only as
\begin{equation}
  L_{img}(\hat{I},I) = \frac{1}{N}\sum_{i}m_{I}(i)\cdot||\hat{I}(i)-I(i)||^2
  \label{eq:eq3}
\end{equation}
\noindent where $m_I(i)$ indicates whether the i-th pixel is masked or not (0 if not masked, 1 otherwise). $N=\sum_{i}m_{I}(i)$ denotes the total number of masked pixels from the input image.

\subsubsection{Total loss.}
A final loss is calculated as the sum of the disparity loss $L_{disp}$ and the image reconstruction loss $L_{img}$.
As random masking is applied to both the left and right images in the student network, $L_{img}$ is computed for both left and right images.
The total loss is defined as $L_{total} = L_{disp} + L_{img}(\hat{I_l},I_l)+L_{img}(\hat{I_r},I_r)$.

\section{Experiments}


\subsection{Implementation Details}

We fine-tuned with pre-trained weights from CroCo-Stereo~\cite{croco2}, adhering to its implementation settings. We evaluated our method on KITTI 2015~\cite{kitti15}, and ETH3D~\cite{eth3d}. The final performance on the KITTI 2015 benchmark in Table~\ref{table:15test} and the ETH3D benchmark in Table~\ref{tab:eth3d} are shown. Further details regarding the training datasets and the specific splits can be found in the supplementary material.

\subsubsection{Pre-training.}
Pre-trained weights from CroCo-Stereo were fine-tuned for our experimentation purposes. CroCo-Stereo pre-trained on various stereo datasets such as CREStereo~\cite{crestereo}, SceneFlow~\cite{sf}, ETH3D~\cite{eth3d}, Booster~\cite{booster}, and Middlebury~\cite{middle}. The pre-trained weights were trained with a batch size of 6 pairs of image crops (704×352) for 32 epochs, utilizing the AdamW optimizer~\cite{adam} with a weight decay rate of 0.05 and a cosine learning rate schedule with a single warm-up epoch, with the learning rate set to $3.10^{-5}$.

\begin{figure*}[t]
\centering
\includegraphics[width=0.85\textwidth]{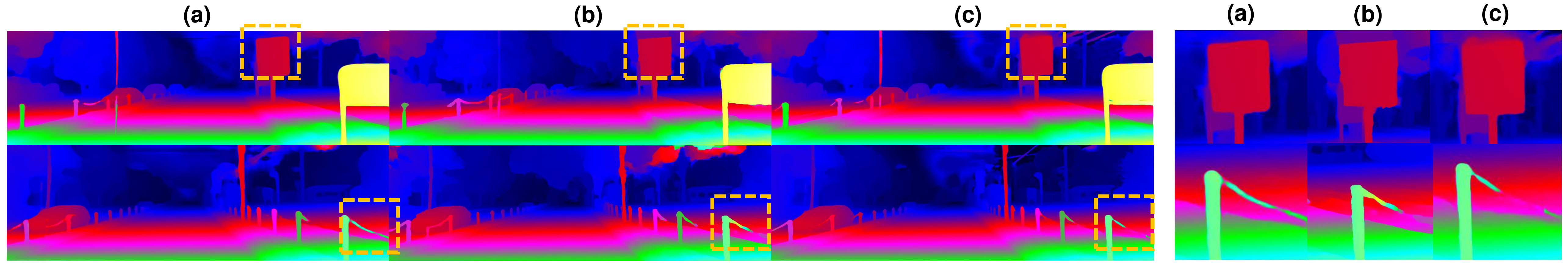} 
\caption{{\bf Qualitative results from the KITTI 2015 Leaderboard}. The left column (a) presents the predicted disparity maps generated by MaDis-Stereo, while the middle column (b) depicts those from IGEV-Stereo. The right column (c) illustrates the results obtained from CroCo-Stereo. The bounding boxes we marked on the artifacts/blur of images are intended for comparison.}
\label{fig:kitti_fig}
\end{figure*}

\subsubsection{Fine-tuning.}
CroCo-Stereo was fine-tuned the pre-trained model on 1216×352 crops from non-masked KITTI 2012~\cite{kitti12} and KITTI 2015~\cite{kitti15} datasets for 20 epochs. Following the most settings of CroCo-Stereo, we used crops of size 1216×352 for fine-tuning, with a learning rate of $3.10^{-5}$. Our MaDis-Stereo adopted the ViT-Base model as its backbone, in contrast to CroCo-Stereo, used the ViT-Large backbone for fine-tuning. Additionally, unlike the original CroCo-Stereo approach, we extended the training of MaDis-Stereo to 100 epochs to ensure effective image reconstruction. For a fair comparison, we compared our results with fine-tuning the CroCo-Stereo for also 100 epochs.

\setlength{\tabcolsep}{5pt}
\begin{table}[t]
\begin{center}
\fontsize{7.5}{8}\selectfont
\caption{ {\bf Results on KITTI 2015 Leaderboard}. We achieved a state-of-the-art result in both the D1-fg and D1-all metrics.}
\vspace{-0.2cm}
\label{table:15test}
\begin{tabular}{llll}
\hline\noalign{\smallskip}
Method & D1-bg($\downarrow$) & D1-fg($\downarrow$) & D1-all($\downarrow$) \\
\noalign{\smallskip}
\hline
\noalign{\smallskip}
HitNet~\cite{hit} & 1.74 & 3.20 & 1.98 \\
PCWNet~\cite{pcw} & \bf{1.37} & 3.16 & 1.67 \\
CREStereo~\cite{crestereo} & 1.45 & 2.86 & 1.69 \\
GMStereo~\cite{gms} & 1.49 & 3.14 & 1.77 \\
IGEV-Stereo~\cite{gev} & 1.38 & 2.67 & 1.59 \\
CroCo-Stereo~\cite{croco2} & 1.38 & 2.65 & 1.59 \\
\rowcolor{gray!20} \bf{MaDis-Stereo} & 1.42 & \bf{2.31} & \bf{1.57}\\
\hline
\end{tabular}
\end{center}
\vspace{-0.3cm}
\end{table}
\setlength{\tabcolsep}{1.4pt}

\setlength{\tabcolsep}{5.5pt}
\begin{table}[t]
\begin{center}
\fontsize{7.5}{8}\selectfont
\caption{{\bf Ablation Study on Masking Ratio $r$}.
We observe the influence of the masking ratio for MaDis-Stereo on KITTI 2015 validation errors. We empirically found that 40\% yielded the best performance across all six metrics.}
\vspace{-0.2cm}
\label{table:ratio}
\begin{tabular}{llllllll}
\hline\noalign{\smallskip}
$r(\%)$ & avgerr & rmse & bad@0.5 & bad@1.0 & bad@2.0 & bad@3.0\\
\noalign{\smallskip}
\hline
\noalign{\smallskip}
10 & 0.503 & 1.472 & 26.557 & 8.546 & 2.789 & 1.008 \\
20 & 0.508 & 1.425 & 27.257 & 8.820 & 2.677 & 1.035 \\
30 & 0.498 & 1.422 & 26.259 & 8.843 & 2.730 & 1.046 \\
\rowcolor{gray!20}\textbf{40} & \textbf{0.469} & \textbf{1.339} & \textbf{25.271} & \textbf{8.113} & \textbf{2.380} & \textbf{0.855} \\
50 & 0.501 & 1.384 & 27.807 & 8.955 & 2.470 & 0.986 \\
60 & 0.515 & 1.397 & 28.277 & 9.399 & 3.125 & 1.109 \\
70 & 0.529 & 1.418 & 29.683 & 9.889 & 3.443 & 1.103 \\
80 & 0.564 & 1.423 & 32.632 & 11.181 & 3.538 & 1.348 \\
90 & 0.793 & 1.692 & 46.985 & 22.279 & 6.258 & 3.070 \\
\hline
\end{tabular}
\end{center}
\vspace{-0.3cm}
\end{table}
\setlength{\tabcolsep}{1.4pt}

\begin{figure}[t]
\centering
\includegraphics[width=0.95\columnwidth]{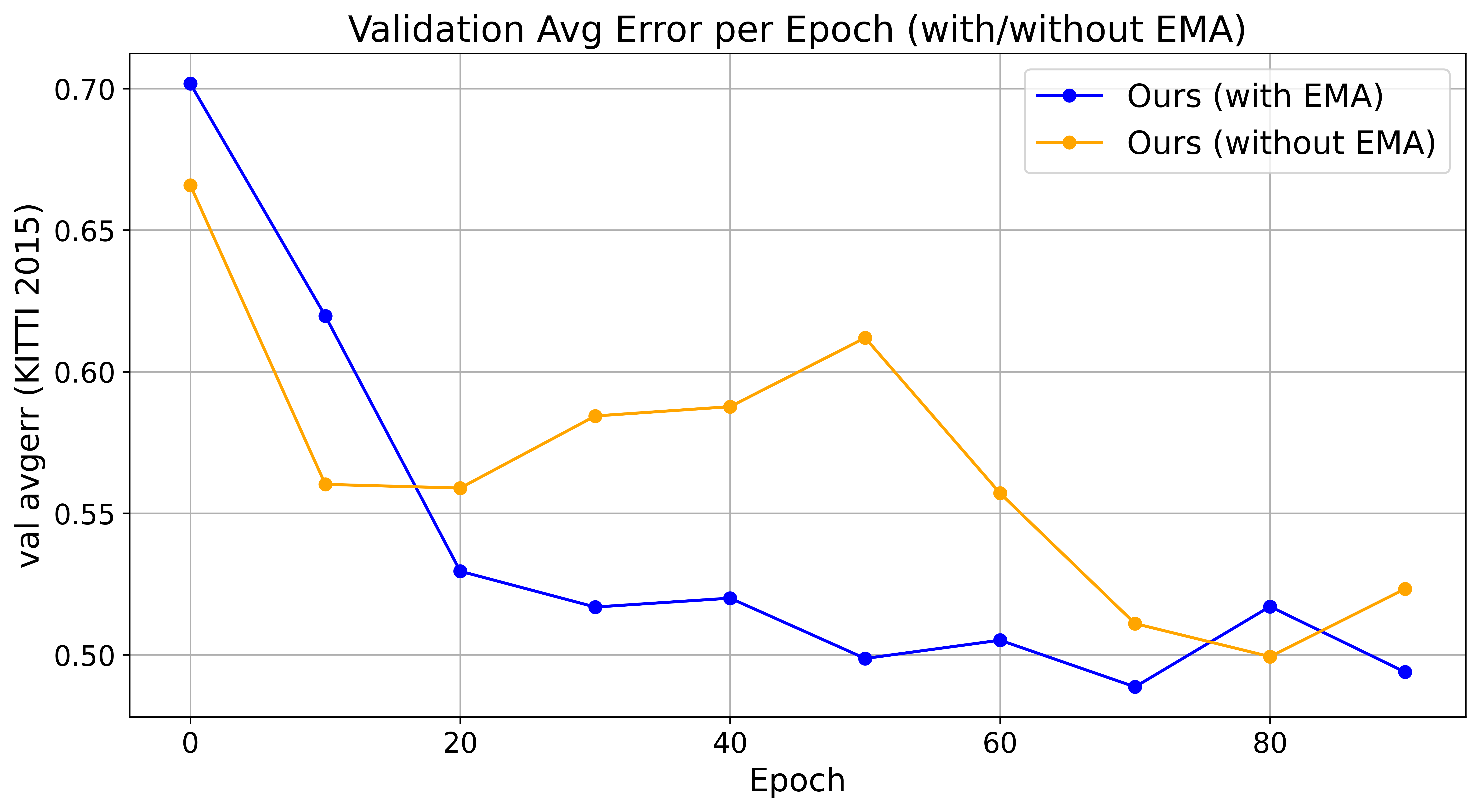} 
\vspace{-5pt}
\caption{{\bf Ablation Study on Exponential Moving Average (EMA)-updated Teacher on KITTI 2015}. 
We observed that the EMA structure stabilizes the training of stereo depth estimation networks, where masking is applied to inputs.}
\label{fig:ema_avgerr}
\vspace{-0.2cm}
\end{figure}

\subsection{Evaluation}

\begin{table*}[t]
\setlength{\tabcolsep}{10.5pt}
  \centering
  \fontsize{8}{8}\selectfont
    \caption{{\bf Results on ETH3D (Low-res two-view benchmark) Leaderboard}. We observed better results than baseline CroCo-Stereo in overall metrics and achieved state-of-the-art performance in bad@0.5, bad@1.0, and bad@4.0.}
  \begin{tabular}{lcccccccccccc}
    \toprule
        \multirow{2}{*}{Method}  & \multicolumn{2}{c}{bad@0.5\((\)\%\()\)$\downarrow$} && \multicolumn{2}{c}{bad@1.0\((\)\%\()\)$\downarrow$} && \multicolumn{2}{c}{bad@4.0\((\)\%\()\)$\downarrow$} && \multicolumn{2}{c}{avg err\((\)px\()\)$\downarrow$}\\
    &  noc & all && noc & all && noc & all && noc & all\\
    \midrule
    HITNet~\cite{hit} & 7.89 & 8.41 && 2.79 & 3.11 && 0.19 & 0.31 && 0.20 & 0.22\\
    RAFT-Stereo~\cite{raftnet} & 7.04 & 7.33 && 2.44 & 2.60 && 0.15 & 0.22 && 0.18 & 0.19\\
    CREStereo~\cite{crestereo} & 3.58 & 3.75 && 0.98 & 1.09 && 0.10 & 0.12 && \textbf{0.13} & \textbf{0.14} \\
    GMStereo~\cite{gms} & 5.94 & 6.44 && 1.83 & 2.07 && 0.08 & 0.14 && 0.19 & 0.21\\
    IGEV-Stereo~\cite{gev} & 3.52 & 3.97 && 1.12 & 1.51 && 0.11 & 0.41 && 0.14 & 0.20\\
    CroCo-Stereo~\cite{croco2} & 3.27 & 3.51 && 0.99 & 1.14 && 0.13 & 0.18 && 0.14 & 0.15 \\
    \rowcolor{gray!20}\textbf{MaDis-Stereo} & \textbf{2.73} & \textbf{2.96} && \textbf{0.80} & \textbf{0.96} && \textbf{0.06} & \textbf{0.11} && 0.14 & 0.15	
\\
    \bottomrule
  \end{tabular}
  \label{tab:eth3d}
  \vspace{-0.3cm}
\end{table*}

We evaluated our model on KITTI 2015~\cite{kitti15} and ETH3D~\cite{eth3d} two-view stereo. We experimentally confirmed that our model is most effective when conducting image reconstruction concurrently, particularly at 100 epochs. Similarly, we compared the results of CroCo-Stereo trained for up to 100 epochs with ViT-Base backbone. Table~\ref{table:15test} for MaDis-Stereo results on KITTI 2015 leaderboard. MaDis-Stereo achieves the best result on the main D1-all metrics with 1.57, with the best value also on D1-fg pixels with 2.31. Furthermore, Table~\ref{tab:eth3d} shows the results on the ETH3D leaderboard, where our model demonstrates state-of-the-art performance. 

\begin{table}[t]
\setlength{\tabcolsep}{4.5pt}
\begin{center}
\fontsize{8}{8}\selectfont
\caption{{\bf Ablation Study on the impact of the EMA-Teacher}. 
We determined that the EMA-Teacher network used in the training process of stereo matching networks on the KITTI 2015 dataset enhances the final performance.}
\vspace{-0.2cm}
\label{table:ema}
\begin{tabular}{lllllllll}
\hline\noalign{\smallskip}
EMA & avgerr & rmse & bad@0.5 & bad@1.0 & bad@2.0 & bad@3.0\\
\noalign{\smallskip}
\hline
\noalign{\smallskip}
$\times$ & 0.489 & 1.379 & 26.102 & 8.669 & 2.370 & 0.981 \\
\rowcolor{gray!20}\textbf{O} & 0.469 & 1.339 & 25.271 & 8.113 & 2.380 & 0.855 \\
\hline
\end{tabular}
\end{center}
\vspace{-0.2cm}
\end{table}

\setlength{\tabcolsep}{4.5pt}
\begin{table}[t]
\scriptsize
\begin{center}
\caption{{\bf Ablation Study on Loss Weight of $L_{disp}$}.
We examine the impact of the $L_{disp}$ for MaDis-Stereo on KITTI 2015  validation errors. As the weight of $L_{disp}$ increases, our model achieves better performance (best at loss weight 1), indicating that the $L_{disp}$ effectively serves its guidance role.}
\vspace{-0.2cm}
\label{table:loss}
\begin{tabular}{lllllllll}
\hline\noalign{\smallskip}
$L_{img} / L_{disp}$ & avgerr & RMSE & bad@0.5 & bad@1.0 & bad@2.0 & bad@3.0\\
\noalign{\smallskip}
\hline
\noalign{\smallskip}
CroCo-Stereo & 0.489 & 1.393 & 27.204 & 8.640 & 2.412 & 0.889 \\
(Ours) 1/0.1 & 0.484 & 1.343 & 26.473 & 8.378 & 2.474 & 0.930 \\
(Ours) 1/0.3 & 0.482 & 1.332 & 26.337 & 8.302 & 2.500 & 0.909 \\
(Ours) 1/0.5 & 0.482 & 1.346 & 25.381 & 8.286 & 2.562 & 0.909 \\
(Ours) 1/0.7 & 0.480 & 1.339 & 26.112 & 8.200 & 2.632 & 0.917 \\
\rowcolor{gray!20}\textbf{(Ours) 1/1} & 0.469 & 1.339 & 25.271 & 8.113 & 2.380 & 0.855 \\
\hline
\end{tabular}
\end{center}
\vspace{-0.3cm}
\end{table}
\setlength{\tabcolsep}{1.4pt}

\subsection{Ablation Study} \label{Sec:ablation}
Our ablations were performed on the KITTI 2015~\cite{kitti15} dataset  for stereo matching. Here, we conducted three ablation studies in this section. Firstly, we analyzed the impact of the masking ratio $r$ on MaDis-Stereo. Secondly, we compared the results with/without the EMA structure to assess its influence on the model's stability. Lastly, we explored the effect of utilizing pseudo disparity maps generated by the EMA teacher network to guide the student network.

\subsubsection{Masking ratio}
We measure the impact of variation in the masking ratio in Table~\ref{table:ratio}. The masking ratio is one of the crucial factors influencing the performance of the MIM network. The model learns a locality inductive bias for the missing parts in the process of predicting the masked regions of the image, which is determined by the masking ratio. However, a higher masking ratio reduces the visible image tokens available for reconstruction and makes it a challenging task. To mitigate the difficulty of the applied MIM strategy, we use a lower masking ratio than what is typically used in the MIM pre-training process. We observed that masking ratio $r = 0.4$ is optimized for MaDis-Stereo.

\subsubsection{EMA structure}
Fig~\ref{fig:ema_avgerr} illustrates the outcomes achieved with/without the utilization of the EMA structure and each validation error results are described in Table~\ref{table:ema}. To ensure more stable learning, we build our model with an EMA structure. Following the standard EMA model, our MaDis-Stereo consists of the teacher and student network. The parameters of the student network are EMA to the parameters of the teacher network. The model without an EMA structure implies that solely the student network is employed for training. Our observations indicate that the MaDis-Stereo with the EMA structure exhibits much more stable learning than the model without it.

\subsubsection{Effects of using pseudo disparity map as supplement guidance}
Performing image reconstruction and depth estimation simultaneously presents challenges. To tackle this issue, we utilize pseudo-disparity maps generated by an EMA teacher network as guidance to the student network during training. Table~\ref{table:loss} shows the impact of varying the weight of $L_{disp}$, which utilizes the disparity map as guidance.

\section{Conclusion}
We investigate the limitations of Transformer-based fine-tuning models for stereo depth estimation tasks and introduce a novel framework that integrates supervised and self-supervised learning approaches, diverging from conventional supervised methods. Our focus lies in leveraging the masking-and-reconstruction approach to enhance the inductive locality bias essential for scenarios with limited data, thereby addressing the bias deficiency in Transformers. Experimentally, we demonstrate the beneficial impact of learning local image representations during fine-tuning on a stereo depth estimation network.

\bibliography{aaai25}

\end{document}